\documentclass[journal]{IEEEtran}

\ifCLASSINFOpdf
  \usepackage[pdftex]{graphicx}

  \graphicspath{{./figures}}

\else
  \usepackage[dvips]{graphicx}

  \graphicspath{{./figures/}}
\fi

\usepackage{gensymb}
\usepackage{color}
\usepackage{amsmath}
\usepackage{amsfonts}
\usepackage{amssymb}

\newcommand{\ra}[1]{\renewcommand{\arraystretch}{#1}}
\usepackage{setspace}
\usepackage{adjustbox}
\usepackage{booktabs}

\usepackage{footnote}
\makesavenoteenv{table}
\makesavenoteenv{tabular}
\RequirePackage{pdfpages}

\begin{document}

\begin{titlepage}
\includepdf{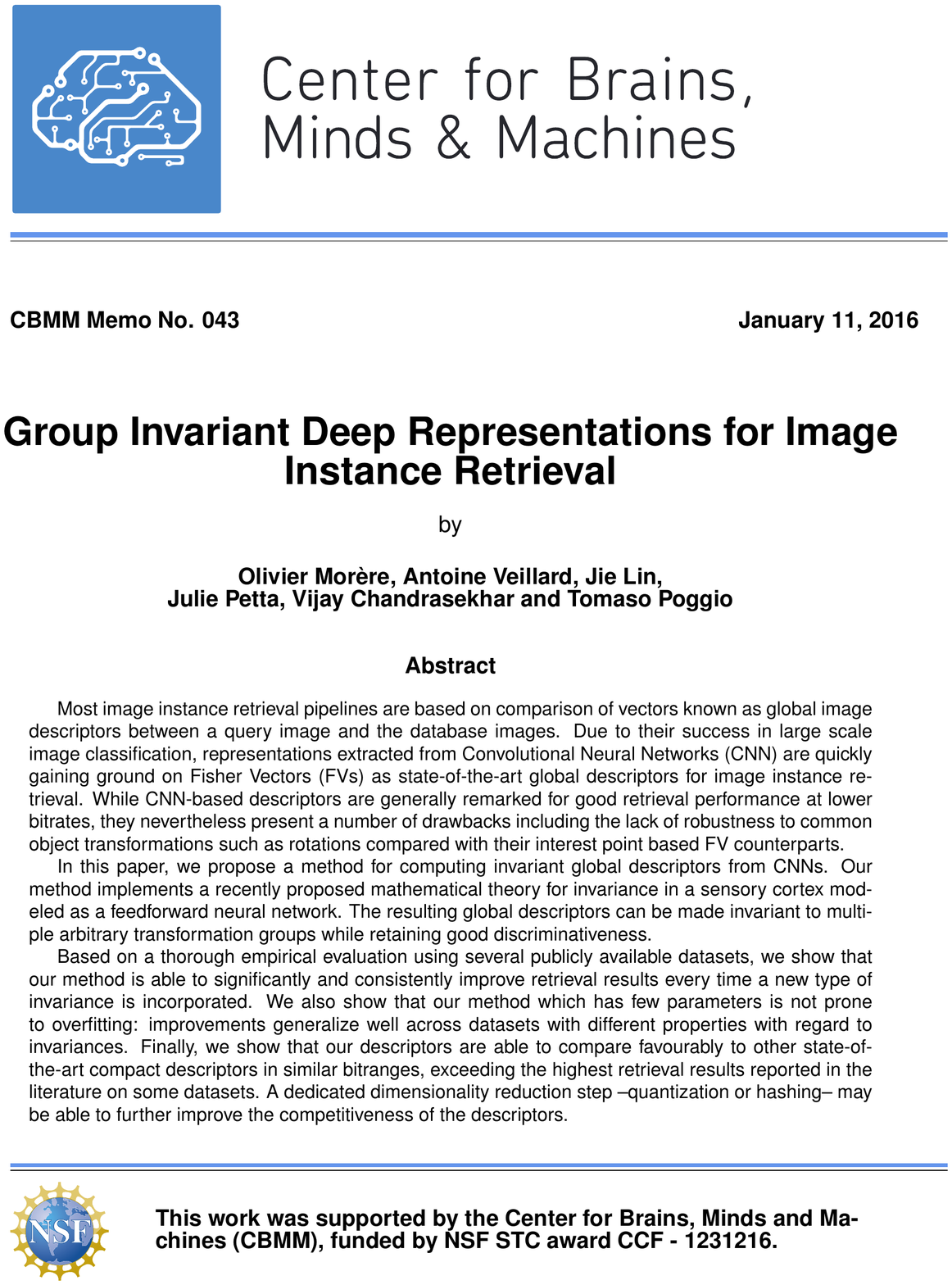}
\end{titlepage}

\title{Group Invariant Deep Representations for Image Instance Retrieval}

\author{Olivier Mor\`{e}re{$^{*,1,2}$}, Antoine Veillard{$^{*,1}$}, Jie Lin{$^{2}$}\\
Julie Petta{$^{3}$, Vijay Chandrasekhar{$^{2}$}, Tomaso Poggio{$^{4}$}}
\thanks{{$^{*}$} O. Mor\`ere and A. Veillard contributed equally to this work.}
\thanks{{$^{1}$} Universit\'e Pierre et Marie Curie}
\thanks{{$^{2}$} A*STAR Institute for Infocomm Research}
\thanks{{$^{3}$} CentraleSup\'{e}lec}
\thanks{{$^{4}$} CBMM, LCSL, IIT and MIT}
}

\maketitle

\begin{abstract}

Most image instance retrieval pipelines are based on comparison of vectors known as global image descriptors between a query image and the database images.
Due to their success in large scale image classification, representations extracted from Convolutional Neural Networks (CNN) are quickly gaining ground on Fisher Vectors (FVs) as state-of-the-art global descriptors for image instance retrieval.
While CNN-based descriptors are generally remarked for good retrieval performance at lower bitrates, they nevertheless present a number of drawbacks including the lack of robustness to common object transformations such as rotations compared with their interest point based FV counterparts.

In this paper, we propose a method for computing invariant global descriptors from CNNs.
Our method implements a recently proposed mathematical theory for invariance in a sensory cortex modeled as a feedforward neural network.
The resulting global descriptors can be made invariant to multiple arbitrary transformation groups while retaining good discriminativeness.

Based on a thorough empirical evaluation using several publicly available datasets, we show that our method is able to significantly and consistently improve retrieval results every time a new type of invariance is incorporated.
We also show that our method which has few parameters is not prone to overfitting: improvements generalize well across datasets with different properties with regard to invariances.
Finally, we show that our descriptors are able to compare favourably to other state-of-the-art compact descriptors in similar bitranges, exceeding the highest retrieval results reported in the literature on some datasets.
A dedicated dimensionality reduction step --quantization or hashing-- may be able to further improve the competitiveness of the descriptors.

\end{abstract}


\section{Intoduction}
\label{sec:introduction}

Image instance retrieval is the discovery of images from a database representing the same object or scene as the one depicted in a query image.
The first step of a typical retrieval pipeline starts with the comparison of vectors representing the image contents known as {\it global image descriptors}.
Good quality descriptors is key to achieving good retrieval performances.

Deep learning neural networks are fast becoming the dominant approach for image classification due to their remarkable performance for large scale image classification~\cite{AlexNet,Simonyan2014}. 
In their recent work, Babenko et al.~\cite{Yandex} propose using representations extracted from Convolutional Neural Nets (CNN) as a global descriptor for image retrieval, and show promising initial results for the approach.
In our recent work~\cite{CompactGlobal, unsupervisedtriplethashing}, we also show how stacked Restricted Boltzmann Machines (RBM) and supervised fine-tuning can be used for generating extremely compact hashes from global descriptors obtained from CNNs for large scale image-retrieval.

While CNN based descriptors are progressively replacing Fisher Vectors (FV)~\cite{Perronnin_CVPR_10} as state-of-the-art descriptors for image instance retrieval, we have shown in our recent work thoroughly comparing both types of descriptors~\cite{practicalguide2015} that the use of CNNs still presents a number of significant drawbacks compared with FVs.
One of them is the lack of invariance to transformations of the input image such as rotations: the performance of CNN descriptors quickly degrade when the objects in the query and the database image are rotated differently.

In this paper, we propose a method to produce global image descriptors from CNNs which are both compact and robust to such transformations.
Our method is inspired from a recent invariance theory (subsequently referred to as \emph{i-theory}) for information processing in sensory cortex~\cite{itheory1, itheory2, itheory3}.
The theory is an information processing model explaining how feedforward information processing in a sensory cortex can be made robust to various types of signal distorsions.
In particular, it provides a practical and mathematically proven way for computing invariant object representations with feedforward neural networks.

After showing that CNNs are compatible with the \emph{i-theory}, we propose a simple and practical way to apply the theory to the construction of global image descriptors which are robust to various types of transformations of the input image at the same time.
Through a thorough empirical evaluation based on multiple publicly available datasets, we show that our method is able to significantly consistently improve retrieval results while keeping dimensionality low.
Rotations, translations and scale changes are studied in the scope of this paper but our approach is extensible to other types of transformations.

\begin{figure*}
\includegraphics[width=\textwidth]{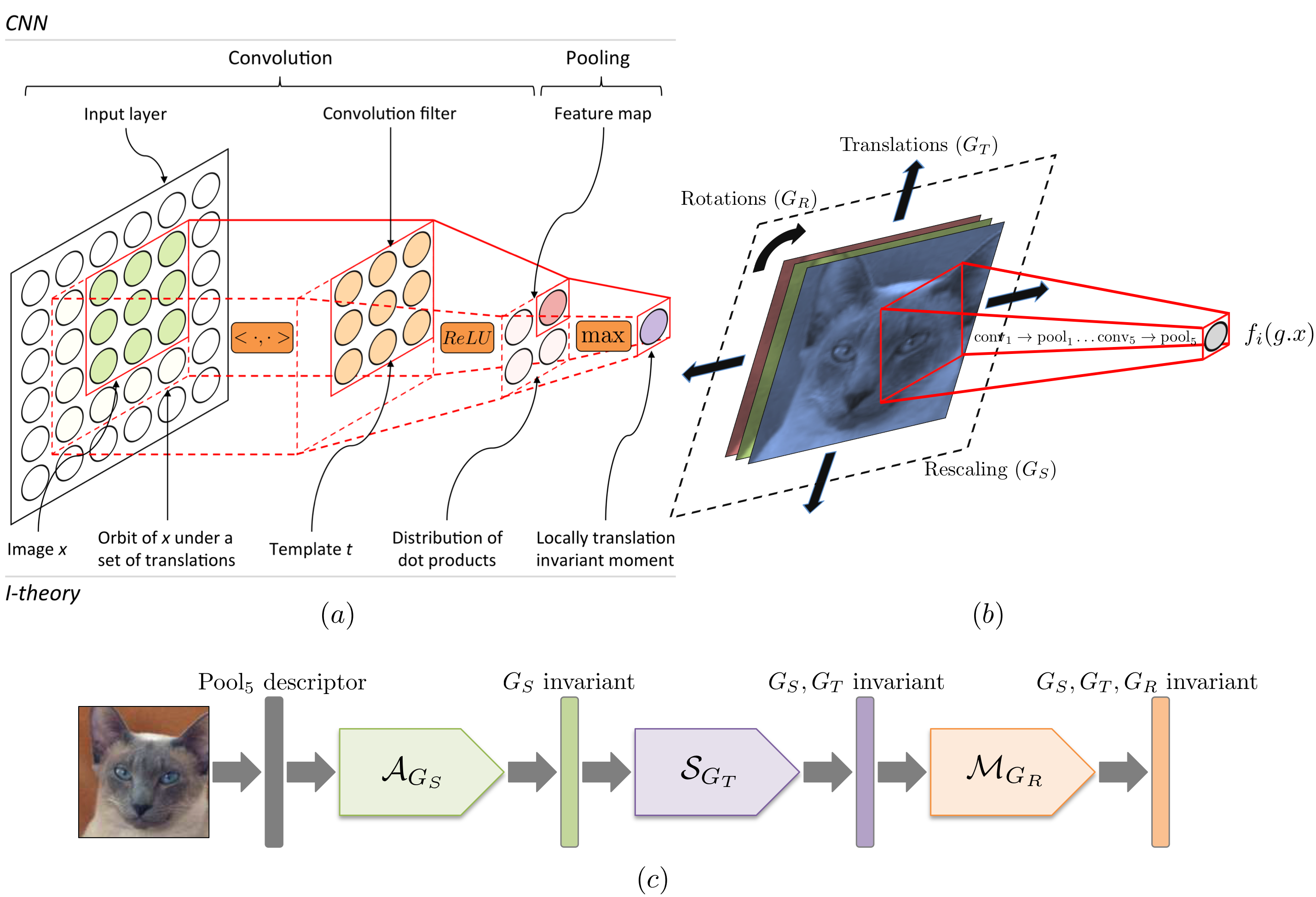}
\caption{(a) A single convolution-pooling operation from a CNN schematized for a single input layer and single output neuron.
The parallel with \emph{i-theory} shows that the universal building block of CNNs is compatible with the incorporation of invariance to local translations of the input according to the theory.
The network architecture is responsible for the invariance properties while back-propagation provides a practical way to learn the templates from data.
(b) A specific succession of convolution and pooling operations learnt by the CNN (depicted in red) computes the \emph{pool5} feature $f_i$ for each feature map $i$ from the RGB image data.
A number of transformations $g$ can be applied to the input $x$ in order to vary the response $f_i(g.x)$.
(c) Our method takes inspiration from the \emph{i-theory} to create compact and robust global image descriptors from CNNs.  
Starting with raw \emph{pool5} descriptors, it can be used to stack-up an arbitrary number of transformation group invariances while keeping the dimensionality under control.
The particular sequence of transformation groups and statistical moments represented on the diagram was shown to produce the best performing hashes on average in our study but other arbitrary combinations are also able to improve retrieval results.
}
\label{fig:method}
\end{figure*}

\section{Related Work}
\label{sec:related}

Since the winning submission of Krizhevsky et al. in the ImageNet 2012 challenge~\cite{AlexNet}, deep CNNs are now considered as the mainstream go-to approach for large-scale image classification.
They are also known to achieve state-of-the-art results with many other visual recognition tasks such as face recognition~\cite{deepface,deepid}, pedestrian detection~\cite{deeppedestrian} and pose estimation~\cite{deeppose}.

CNN began to be applied to the instance retrieval problem as well, although there is comparatively less work on CNN-based descriptors for instance retrieval compared to large-scale image classification.
Razavian et al.~\cite{CNNOffTheShelf} evaluate the performance of CNN model of~\cite{AlexNet} on a wide range of tasks including instance retrieval, and show initial promising results.
Babenko et al.~\cite{Yandex} show that a pre-trained CNN can be fine tuned with domain specific data (objects, scenes, etc.) to improve retrieval performance on relevant data sets.
The authors also show that the CNN representations can be compressed more effectively than their FV counterparts for large-scale instance retrieval.
In~\cite{CompactGlobal, unsupervisedtriplethashing}, we show how sparse high-dimensional CNN representations can be hashed to very compact representations (32-1024 bits) for large scale image retrieval with little loss in matching performance.

Meanwhile, we showed in our recent evaluation work~\cite{practicalguide2015} that CNN-based descriptors still suffer from a number of drawbacks including the lack of robustness to certain transformations of the input data.
The \emph{i-theory} proposed as the guideline method in this paper was recently used to successfully compute robust representations for face recognition \cite{poggioface} and music classification \cite{poggiomusic}.

The contributions of our work can be summarized as follows.
\begin{itemize}
\item A method based on \emph{i-theory} for creating robust and compact global image descriptors from CNNs.
\item The ability to iteratively incorporate different group invariances, each new addition leading to consistent and significant improvements in retrieval results.
\item A set of highly competitive global descriptors for image instance retrieval compared with other state-of-the-art compact descriptors at similar bitrates.
\item A low risk of overfitting: few parameters and many reasonable settings which can generalize well across all datasets.
\item A thorough empirical study based on several publicly available datasets.
\end{itemize}

\section{Invariant Global Image Descriptors}
\label{sec:method}

\subsection{I-theory in an Nutshell}
\label{sec:method1}

Many common classes of image transformations such as rotations, translations and scale changes can be modeled by the action of a group $G$ over the set $E$ of images.
Let $x \in E$ and a group $G$ of transformations acting over $E$ with group action $G \times E \rightarrow E$ denoted with a dot ($.$).
The orbit of $x$ by $G$ is the subset of $E$ defined as $O_x = \{ g.x \in E | g \in G \}$.
It can be easily shown that $O_x$ is globally invariant to the action of any element of $G$ and thus any descriptor computed directly from $O_x$ would be globally invariant to $G$.

The \emph{i-theory} predicts that an invariant descriptor for a given object $x \in E$ is computed in relation with a predefined template $t \in E$ from the distribution of the dot products $D_{x,t} = \{< g.x , t >  \in \mathbb{R} | g \in G \} = \{< x , g.t > \in \mathbb{R} | g \in G \}$ over the orbit. 
One may note that the transformation can be applied either on the image or the template indifferently.
The proposed invariant descriptor extracted from the pipeline should be a histogram representation of the distribution with a specific bin configuration.
Such a representation is mathematically proven to have proper invariance and selectivity properties provided that the group is compact or at least locally compact \cite{itheory1}.

In practice, while a compact group (e.g. rotations) or locally-compact group (e.g. translations, scale changes) is required for the theory to be mathematically provable, the authos of \cite{itheory1} suggest that the theory extends well (with approximate invariance) to non-locally compact groups and even to continuous non-group transformations (e.g. out-of-plane rotations, elastic deformations) provided that proper class-specific templates can be provided.
Recent work on face verification \cite{poggioface} and music classification \cite{poggiomusic} apply the theory to non-compact groups with good results.
Additionally, the histograms can also be effectively replaced by statistical moments (e.g. mean, min, max, standard deviation, etc.).

\subsection{CNNs are i-theory Compliant Networks}
\label{sec:method2}

All popular CNN architectures designed for image classification such as \emph{AlexNet}~\cite{AlexNet} and \emph{OxfordNet}~\cite{Simonyan2014} share a common building block: a succession of convolution-pooling operations designed to model increasingly high-level visual representations of the data.
The highest level visual features may then be fed into fully connected layers acting as a classifiers.

As shown in detail on Figure~\ref{fig:method}~(a), the succession of convolution and pooling operations in a typical CNN is in fact a way to incorporate local translation invariance strictly compliant with the framework proposed by the \emph{i-theory}.
The network architecture provides the robustness such as predicted by the invariance theory while training via back propagation ensures a proper choice of templates.
In general, multiple convolution-pooling steps are applied (5 times in both \emph{AlexNet} and \emph{OxfordNet}) resulting in increased robustness and higher level templates.
Note that the iterative composition of local translation invariance approximately translates into robustness to local elastic distortions for the features at the \emph{pool5} layer.

In this study, instead of the popular first fully-connected layer (\emph{fc6}) which is on average the best single CNN layer to use as a global out-of-the-box descriptor for image retrieval \cite{practicalguide2015}, we decide to use the locally invariant \emph{pool5} as a starting representation for our own global descriptors and further enhance their robustness to selected transformation groups in a way inspired from \emph{i-theory}.

\subsection{Transformation Invariant CNN Descriptors}
\label{sec:method3}

For every feature map $i$ of the \emph{pool5} layer ($0 \leq i < 512$ in the case of the presently used {\it OxfordNet}), we denote $f_i(x)$ the corresponding feature obtained from the RGB image data through a succession of convolution-pooling operations.
Note that the transformation $f_i$ is non-linear and thus not strictly a mathematical dot product with a template but can still be viewed as an inner product.

In this study, we propose to further improve the invariance of \emph{pool5} CNN descriptors by incorporating global invariance to several transformation groups.
The specific transformation groups considered in this study are translations $G_T$, rotations $G_R$ and scale changes $G_S$.
As shown on Figure~\ref{fig:method}~(b), transformations $g$ are applied on the input image $x$ varying the output of the \emph{pool5} feature $f_i(g.x)$ accordingly.

The invariant statistical moments computed from the distributions $\{ f_i(g.x) | g \in G \}$ with $G \in \{ G_T, G_R, G_S \}$ are averages, maxima and standard deviations, respectively:
\begin{align} 
&{\mathcal{A}}_{G,i}(x) = \displaystyle \frac{1}{\int_G dg}\int_G f_i(g.x) dg \label{eqn1}\\
&{\mathcal{M}}_{G,i}(x) = \displaystyle \max_G \left( f_i(g.x) \right) \label{eqn2}\\
&{\mathcal{S}}_{G,i}(x) = \displaystyle \frac{1}{\int_G dg} \sqrt{ \int_G f_i(g.x)^2 dg - (\int_G f_i(g.x) dg)^2} \label{eqn3}
\end{align}
with corresponding global image descriptors obtained by simply concatenating the moments for the individual features:
\begin{align} 
&{\mathcal{A}}_G(x) = ( {\mathcal{A}}_{G,i}(x) )_{0 \leq i < 512}\\
&{\mathcal{M}}_G(x) = ( {\mathcal{M}}_{G,i}(x) )_{0 \leq i < 512}\\
&{\mathcal{S}}_G(x) = ( {\mathcal{S}}_{G,i}(x) )_{0 \leq i < 512}
\end{align}
In principle, $G$ is always measurable and of finite measure as required since $G_T$ and $G_S$ must be restricted to compact subsets due to image border effects.

An interesting aspect of the \emph{i-theory} is the possibility in practice to chain multiple types of group invariances one after the other as already demonstrated in~\cite{poggiomusic}.
In this study, we construct descriptors invariant to several transformation groups by successively applying the method to different transformation groups as shown on Figure~\ref{fig:method}~(c).
For instance, following scale invariance with average by translation invariance with standard deviation for feature $i$ would correspond to:
\begin{align} 
&\displaystyle \max_{{g_t} \in G_T} \left( \frac{1}{\int_{g_s \in G_S} dg_s}\int_{g_s \in G_S} f_i(g_tg_s.x) dg_s \right)
\end{align}
One may note that the operations are sometimes commutable (e.g. $\mathcal{A}_G$ and $\mathcal{A}_{G'}$) and sometimes not (e.g. $\mathcal{A}_G$ and $\mathcal{M}_{G'}$) depending on the specific combination of moments.

\section{Image Instance Retrieval}
\label{sec:results}

\subsection{Evaluation Framework}

We evaluate our invariant image descriptors in the context of image instance retrieval.
As our starting representation, we use the \emph{pool5} layer from the 16 layers \emph{OxfordNet} \cite{Simonyan2014} with a total dimensionality of $25088$ organized in $512$ feature maps of size $7 \times 7$.

Similarly to our evaluation work in \cite{practicalguide2015}, the rotated input images are padded with the mean pixel value from the ImageNet data set.
The step size for rotations is 10 degrees yielding 36 rotated images per orbit.
For scale changes, 10 different center crops geometrically spanning from 100\% to 50\% of the total image have been taken.
For translations, the entire feature map is used for every feature, resulting in an orbit size of $7 \times 7 = 49$.

We evaluate the performances of the descriptors against four popular data sets: {\it Holidays}, {\it Oxford buildings (Oxbuild)}, {\it UKBench (UKB)} and {\it Graphics}.
The four datasets are chosen for the diversity of data they provide: {\it UKBench} and {\it Graphics} are object-centric featuring close-up shots of objects in indoor environments. 
{\it Holidays} and {\it Oxbuild} are scene-centric datasets consisting primarily of outdoor buildings and scenes.

{\textbf{INRIA Holidays.}
The INRIA Holidays dataset~\cite{Jegou08} consist of personal holiday pictures. 
The dataset includes a large variety of outdoor scene types: natural, man-made, water and fire effects. 
There are 500 queries and 991 database images.
Variations in lighting conditions are rare in this data set as the pictures from the same location are taken at the same time.

\textbf{Oxbuild.}
The Oxford Buildings Dataset~\cite{Philbin07} consists of 5062 images collected from Flickr representing landmark buildings in Oxford. 
The collection has been manually annotated to generate a comprehensive ground truth for 11 different landmarks, each represented by 5 possible queries. 
Note that the set contains 55 queries only. 

{\textbf{UKBench.}}
The University of Kentucky (UKY) data set~\cite{Nister06} consists of 2550 groups of common objects.
There are 4 images representing each.
Only the object of interest present in each image.
Thus, there is no foreground or background clutter within this data set.
All 10200 images are used as queries.

{\textbf{Graphics.}}
The Graphics data set is part of the Stanford Mobile Visual Search data set~\cite{SVMSDataSet}, which notably was used in the MPEG standard: Compact Descriptors for Visual Search (CDVS)~\cite{MPEGDataset2}.
The data set contains different categories of objects like CDs, DVDs, books, software products, business cards, etc.
The query images include foreground and background clutter that would be considered typical in real-world scenarii, e.g., a picture of a CD might contain other CDs in the background.
This data set distinguishes from the other ones as it contains images of rigid objects captured under widely varying lighting conditions, perspective distortion, foreground and background clutter. 
Query images are taken with heterogeneous phone cameras. 
Each query has two relevant images.
There are 500 unique objects, 1500 queries, and 1000 database images.

\subsection{Pairwise Matching Distance}

Image instance retrieval starts with the construction of a list of database images ordered according to their pairwise matching distance with the query image.
With CNN descriptors, the matching distance is strongly affected by commonly encountered image transformations.
As shown in our previous evaluation work~\cite{practicalguide2015}, a rotation of the query image by 10 degrees or more causing a sharp drop in results.
This particular issue is much less pronounced with the popular Fisher vectors, largely due to the use of interest point detectors.

Figure~\ref{fig:dists} provides an insight on how adding different types of invariance with our proposed method will affect the matching distance on different image pairs of matching objects.
With the incorporation of each new transformation group, we notice that the relative reduction in matching distance is the most significant with the image pair which is the most affected by the transformation group. 

\begin{figure}[ht]
\centering
\includegraphics[width=.5\textwidth]{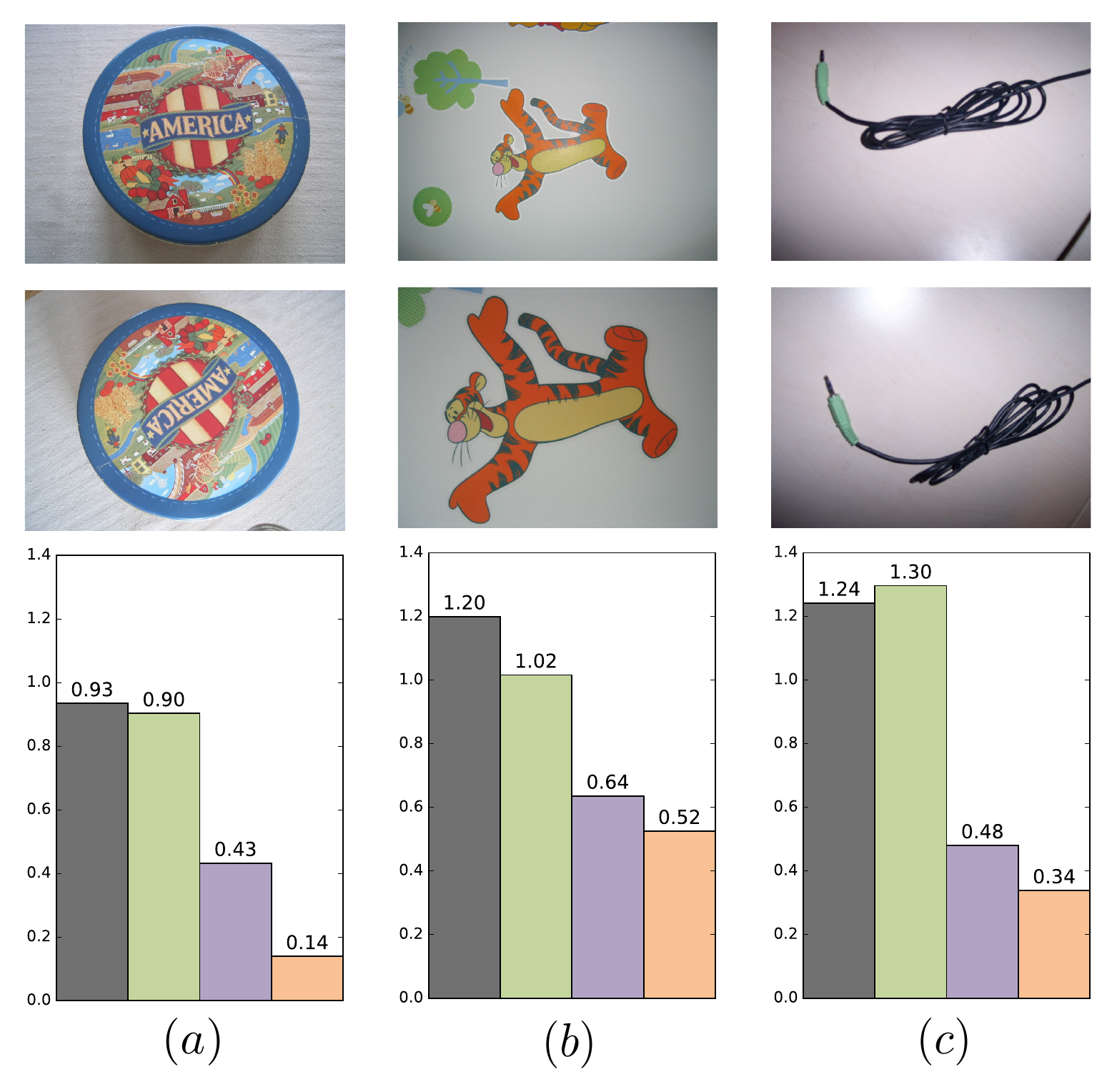}
\begin{tabular}{@{}c@{}cc@{}cc@{}cc@{} c}
\includegraphics[width=.6cm]{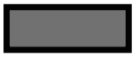}&\emph{pool5}&
\includegraphics[width=.6cm]{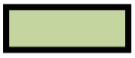}&$\mathcal{A}_{G_S}$&
\includegraphics[width=.6cm]{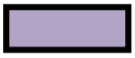}&$\mathcal{A}_{G_S}$-$\mathcal{A}_{G_T}$&
\includegraphics[width=.6cm]{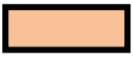} &$\mathcal{A}_{G_S}$-$\mathcal{A}_{G_T}$-$\mathcal{A}_{G_R}$
\end{tabular}
\caption{Distances for 3 matching pairs from \emph{UKBench}. 
For each pair, 4 pairwise distances ($L_2$-normalized) are computed corresponding to the following descriptors: \emph{pool5}, $\mathcal{A}_{G_S}$, $\mathcal{A}_{G_S}$-$\mathcal{A}_{G_T}$ and  $\mathcal{A}_{G_S}$-$\mathcal{A}_{G_T}$-$\mathcal{A}_{G_R}$. 
  Adding scale invariance makes the most difference on (b), translation invariance on (c), and rotation on (a) which is consistent with the scenarii suggested by the images.}
  \label{fig:dists}
\end{figure}

\subsection{Transformations, Order and Moments}

\begin{table}[ht]
\caption{Retrieval results (mAP) for different sequences of transformation groups and moments.
}
\label{tab:res} 
{
\centering
\ra{1.2}
{\footnotesize \singlespacing 
\begin{adjustbox}{max width=\textwidth,center}
\begin{tabular}{@{}lrrrrr@{}}
\toprule
{\sc Sequence} & {\sc Dims} & \multicolumn{4}{c}{\sc Dataset} \\
\cmidrule{3-6}
 & & Oxbuild & Holidays & UKB & Graphics\\
\midrule
\emph{pool5} & 25088 & 0.427 & 0.707 & 0.823(3.105) & 0.315\\
\emph{fc6} & 4096 & 0.461 & 0.782 & 0.910(3.494) & 0.312\\
\midrule
$\mathcal{A}_{G_S}$ & 25088 & 0.430 & 0.716 & 0.828(3.122) & 0.394\\
$\mathcal{A}_{G_T}$ & 512 & 0.477 & 0.800 & 0.924(3.564) & 0.322\\
$\mathcal{A}_{G_R}$ & 25088 & 0.462 & 0.779 & 0.954(3.718) & 0.500\\
$\mathcal{A}_{G_T}$-$\mathcal{A}_{G_R}$ & 512 & 0.418 & 0.796 & 0.955(3.725) & 0.417\\
$\mathcal{A}_{G_T}$-$\mathcal{A}_{G_S}$ & 512 & 0.537 & 0.811 & 0.931(3.605) & 0.430\\
$\mathcal{A}_{G_R}$-$\mathcal{A}_{G_S}$ & 25088 & 0.494 & 0.815 & 0.959(3.752) & 0.552\\
$\mathcal{A}_{G_S}$-$\mathcal{A}_{G_T}$-$\mathcal{A}_{G_R}$ & 512 & 0.484 & 0.833 & 0.971(3.819) & 0.509\\
\midrule
$\mathcal{A}_{G_S}$-$\mathcal{S}_{G_T}$-$\mathcal{M}_{G_R}$ & 512 & \textbf{0.592} & \textbf{0.838} & \textbf{0.975(3.842)} & \textbf{0.589}\\
$\mathcal{A}_{G_S}$-$\mathcal{S}_{G_T}$-$\mathcal{M}_{G_R}$ & 512 {\bf bits} & 0.523 & 0.787 & 0.958(3.741)  & 0.552\\
\bottomrule
\end{tabular}
\end{adjustbox}
}
}
\vspace*{2mm}\\
\footnotesize
Results are computed with the mean average precision (mAP) metric.
For reference, 4$\times$Recall@4 results are also provided for UKBench (between parentheses).
$G_T$, $G_R$, $G_S$ denote the groups of translations, rotations and scale changes respectively.
Note that averages commute with other averages so the sequence order of the composition does not matter when only averages are involved.
Best results are achieved by choosing specific moments.
$\mathcal{A}_{G_S}$-$\mathcal{S}_{G_T}$-$\mathcal{M}_{G_R}$ corresponds to the best average performer for which a binarized version is also given.
\emph{fc6} and \emph{pool5} are provided as a baseline.
\end{table}

\begin{figure}[ht]
  \centering
\includegraphics[width=.42\textwidth]{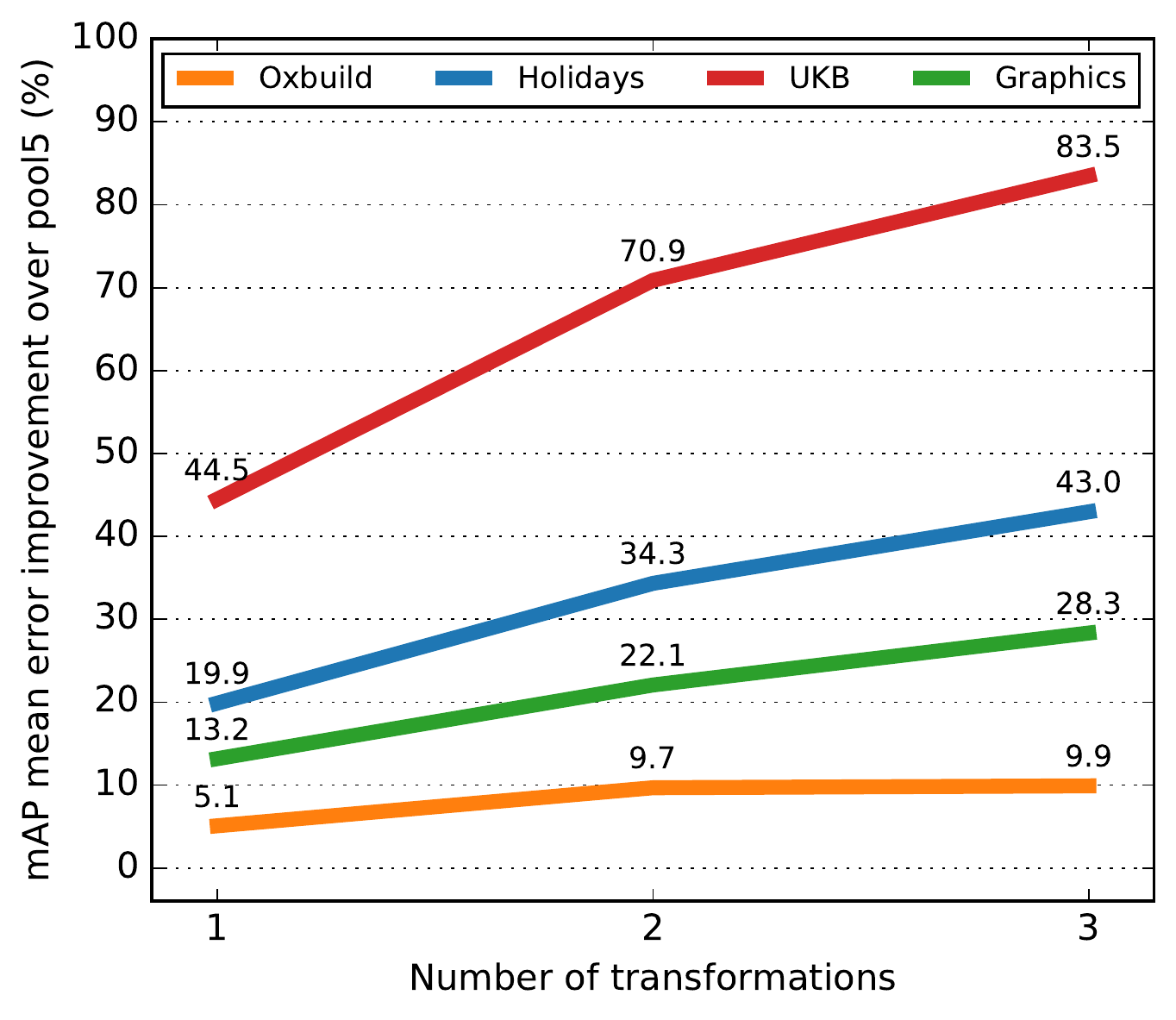}
  \caption{
    Results from Table~\ref{tab:res} for the 7 strategies using averages only (rows 3 to 9) expressed in terms of improvement in mAP over \emph{pool5}, and aggregated by number of invariance groups.
 Improvements range from +5\% on Oxbuild using 1 transformation to +83.5\% on UKBench using 3 transformations.
 On all 4 datasets, results clearly improve with the amount of groups considered.}
  \label{fig:depth}
\end{figure}

\begin{figure}[ht]
\centering
\includegraphics[width=.49\textwidth]{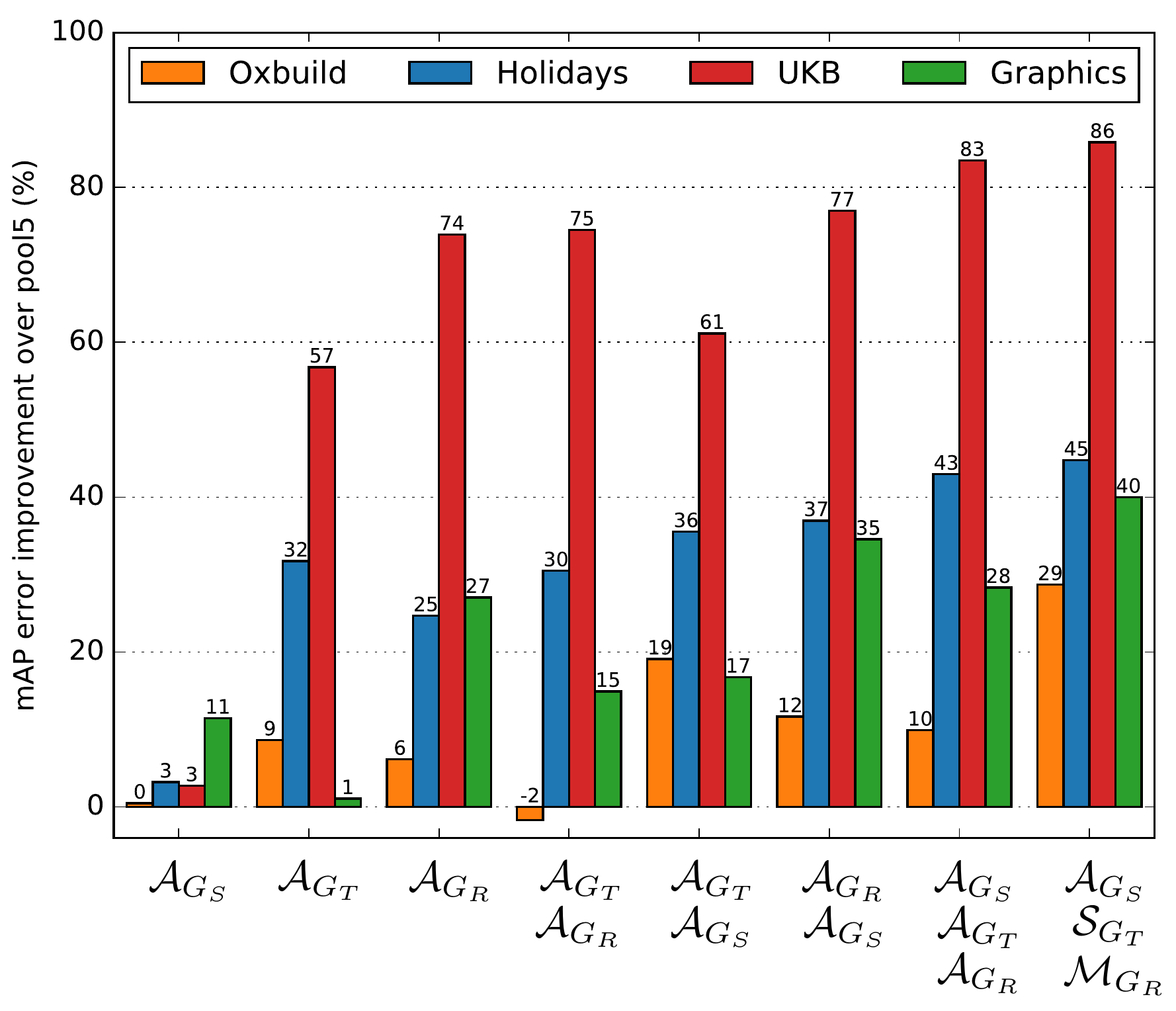}
  \caption{
  Results from Table~\ref{tab:res} expressed in terms of improvement in mAP over \emph{pool5}.
Most strategies yield significant improvements over \emph{pool5} on most datasets.
The average improvement is 68\% for the best strategy. }
  \label{fig:error_improvement}
\end{figure}

Our first set of results summarized in Table~\ref{tab:res} study the effects of incorporating various transformation groups and using different moments.
\emph{Pool5} which is the starting point of our descriptors and \emph{fc6} which is considered the best off-the-shelf descriptor \cite{practicalguide2015,CNNOffTheShelf} are provided as baselines. 
Table~\ref{tab:res} also provides results for all possible combinations of transformation groups for average pooling (order does not matter as averages commute) and for the single best performer which is $\mathcal{A}_{G_S}$-$\mathcal{S}_{G_T}$-$\mathcal{M}_{G_R}$ (order matters).

First, we can immediately point out the high potential of \emph{pool5}.
Although it performs notably worse than \emph{fc6} as-is, a simple average pooling over the space of translations \emph{$\mathcal{A}_{G_T}$} makes it both better and 8 times more compact than \emph{fc6}.
As shown in Figure~\ref{fig:depth}, accuracy increases with the number of transformation groups involved.
On average, single transformation schemes perform 21\% better  compared to \emph{pool5}, 2-transformations schemes perform 34\% better, and the 3-transformations scheme performs 41\% better.

Selecting statistical moments different than averages can further improve the retrieval results.
In Figure~\ref{fig:error_improvement}, we observe that $\mathcal{A}_{G_S}$-$\mathcal{S}_{G_T}$-$\mathcal{M}_{G_R}$ performs roughly 17\% better (average results over all datasets) than $\mathcal{A}_{G_S}$-$\mathcal{A}_{G_T}$-$\mathcal{A}_{G_R}$.
Notably, the best combination corresponds to an increase in the orders of the moments: $\mathcal{A}$ being a first-order moment, $\mathcal{S}$ second order and $\mathcal{M}$ of infinite order.
A different way of stating this fact is that a more invariant representation requires a higher order of pooling.

Overall, $\mathcal{A}_{G_S}$-$\mathcal{S}_{G_T}$-$\mathcal{M}_{G_R}$ improves results over \emph{pool5} by 29\% (Oxbuild) to 86\% (UKBench) with large discrepancies according to the dataset.
Better improvements with UKBench can be explained with the presence of many rotations in the dataset (smaller objects taken under different angles) while Oxbuild consisting mainly of upright buildings is not significantly helped by incorporating rotation invariance.

\subsection{Compact Binary Hashes}

\begin{table}[ht]
\caption{Retrieval performance (mAP) comparing our method to other state-of-the-art methods.}
\label{tab:stateoftheart} 
\centering
\ra{1.2}
{\footnotesize \singlespacing 
\begin{adjustbox}{max width=\textwidth,center}
\begin{tabular}{@{}lrrrr@{}}
\toprule
{\sc Method} & \multicolumn{1}{c}{\sc rep. size} & \multicolumn{3}{c}{\sc Dataset} \\
\cmidrule{3-5}
&  \#dim (size in bits) & Oxbuild & Holidays & UKB\\
\midrule
{\textbf{Our results}} & 512 (512) & 0.523 & \textbf{0.787} & \textbf{0.958}\\
{OxfordNet \cite{sharif2015baseline}} & 256 (1024) & \textbf{0.533} & 0.716 & 0.842\\
{OxfordNet  \cite{sharif2015baseline}} & 256 (256) & 0.436 & 0.578 & 0.693\\
{T-embedding \cite{jegou2014triangulation}} & 256 (2048) & 0.472 & 0.657 & 0.863\\
{T-embedding \cite{jegou2014triangulation}} & 128 (1024) & 0.433 & 0.617 & 0.850\\
{VLAD+CSurf \cite{spyromitros2014comprehensive}} & 128 (1024) & 0.293 & 0.738 & 0.830\\
{mVLAD+Surf  \cite{spyromitros2014comprehensive}} & 128 (1024) & 0.387 & 0.718 & 0.875\\
\bottomrule
\end{tabular}
\end{adjustbox}
}
\vspace*{2mm}\\
\footnotesize
Only methods within a comparable range of bitrates are selected.
\end{table}

As shown in Table~\ref{tab:res}, a simple binarization strategy (thresholding at dataset mean) applied to our best performing descriptor $\mathcal{A}_{G_S}$-$\mathcal{S}_{G_T}$-$\mathcal{M}_{G_R}$ degrades retrieval performance only very marginally and is in fact sufficient to produce a hash that compares favourably with other state-of-the-art methods.
On average over the four datasets, the hash performs 39\% better than \emph{pool5}, while having a 1568 times smaller representation size. Compared to \emph{fc6} feature, the hash performs 26\% better while being 256 times smaller.

As mentioned, the invariant hash also performs well compared to other state-of-the-art.
In Table~\ref{tab:stateoftheart}, we compare our invariant hash against other approaches designed to produce compact representations with comparable bit sizes, 512 being considered fairly compact by current standards \cite{sharif2015baseline}.
In addition, note that the hashing scheme used in this experiment is very naive and better hashes (better retrieval results and/or higher compression rates) could most certainly be obtained by applying more sophisticated hashing methods such as in our previous work~\cite{deephash, unsupervisedtriplethashing} based on stacked RBMs and metric finetuning.

\section{Conclusion}

We proposed a novel method based on \emph{i-theory} for creating robust and compact global image descriptors from CNNs for image instance retrieval.
Through a thorough empirical study, we show that the incorporation of every new group invariance property following the method leads to consistent and significant improvements in retrieval results.

Our method has a number of parameters (sequence of the invariances and choice of statistical moments) but experiments show that many default and reasonable settings produce results which can generalise well across all datasets meaning that the risk of overfitting is low.

This study also confirms the high potential of the feature pyramid (\emph{pool5}) as a starting representation for high-performance compact hashes instead of the more commonly used first fully connected layer (\emph{fc6}).
Our method produces a set of descriptors able to compare favourably with other state-of-the-art compact descriptors at similar bitrates even without dedicated compression/hashing step.

\bibliography{../main}   
\bibliographystyle{IEEEtran}

\end{document}